\pdfoutput=1

\documentclass[final]{article}

\usepackage[final]{acl}

\usepackage{times}
\usepackage{latexsym}

\usepackage[T1]{fontenc}

\usepackage[utf8]{inputenc}

\usepackage{microtype}

\usepackage{inconsolata}
\usepackage{graphicx}
\usepackage[cm,headings]{fullpage}

\title{ \textbf{RFBES at SemEval-2024 Task 8: Investigating Syntactic and Semantic Features for Distinguishing AI-Generated and Human-Written Texts
}}

\author{
Mohammad Heydari Rad\thanks{equal contribution}$^{1}$,
Farhan Farsi\footnotemark[1]$^{1}$,
Shayan Bali\footnotemark[1]$^{1}$, \\
\textbf{Romina Etezadi}$^{2}$ 
\textbf{and Mehrnoush Shamsfard}$^{3}$ \\
$^1$ Computer Engineering Department, Amirkabir University of Technology, Tehran, Iran \\
$^2$ School of Electrical Engineering and Computer Science, University of Ottawa, Ottawa, Canada \\
$^3$ Faculty of Computer Science and Engineering, Shahid Beheshti University, Tehran, Iran \\
{
\small mhrad81@aut.ac.ir,
farhan1379@aut.ac.ir,
shayanbali@aut.ac.ir,
retezadi@uottawa.ca,
m-shams@sbu.ac.ir
}
}

\begin{document}
\maketitle

\vspace{10cm}

\begin{abstract}
Nowadays, the usage of Large Language Models (LLMs) has increased, and LLMs have been used to generate texts in different languages and for different tasks. Additionally, due to the participation of remarkable companies such as Google and OpenAI, LLMs are now more accessible, and people can easily use them. However, an important issue is how we can detect AI-generated texts from human-written ones. In this article, we have investigated the problem of AI-generated text detection from two different aspects: semantics and syntax. Finally, we presented an AI model that can distinguish AI-generated texts from human-written ones with high accuracy on both multilingual and monolingual tasks using the M4 dataset. According to our results, using a semantic approach would be more helpful for detection. However, there is a lot of room for improvement in the syntactic approach, and it would be a good approach for future work.
\end{abstract}

\section{Introduction}
Large Language Models (LLMs) are widely used. They are easily accessible, and people can use them by passing their queries to chatbots to generate their desired texts for several purposes and, more importantly, in different languages. Although LLMs have their own advantages and simplify the text generation process for humans, they have increased concerns about the misuse of this technology for adversarial purposes such as generating hallucinations, misinformation, disinformation, and fake news. Furthermore, improper use of LLMs can cause disruption in students' learning process.

This issue has led to research on detecting AI-generated texts versus human-written ones, and a number of articles have investigated this classification task. However, the main focus of the presented works has mainly been on the semantic aspect of this text classification task. In this article, we have investigated this issue using two different approaches to consider both the semantic and syntactic aspects of texts. To achieve this aspiration, we have developed two different models for both syntactic and semantic-based analysis to apply them to both multilingual and monolingual datasets. In this way, we used the M4 article's dataset \cite{wang2023m4} for both multilingual and monolingual tasks.

For the syntax analysis of this task, we have developed an Attention-based Long Short-Term Memory(LSTM) model to cover the complexities related to long sentences and the relationship between different parts of a sentence, and regarding the semantic analysis of this task, we have developed a transformer-based model.

According to the results, our systems have performed better than M4's provided baseline in the multilingual task, and in the monolingual task, our results are really close to M4's provided baseline. In the end, we have provided our results and compared the results of both these models with each other and with previous works in this area.

\begin{figure*}
    \includegraphics[width=16cm]{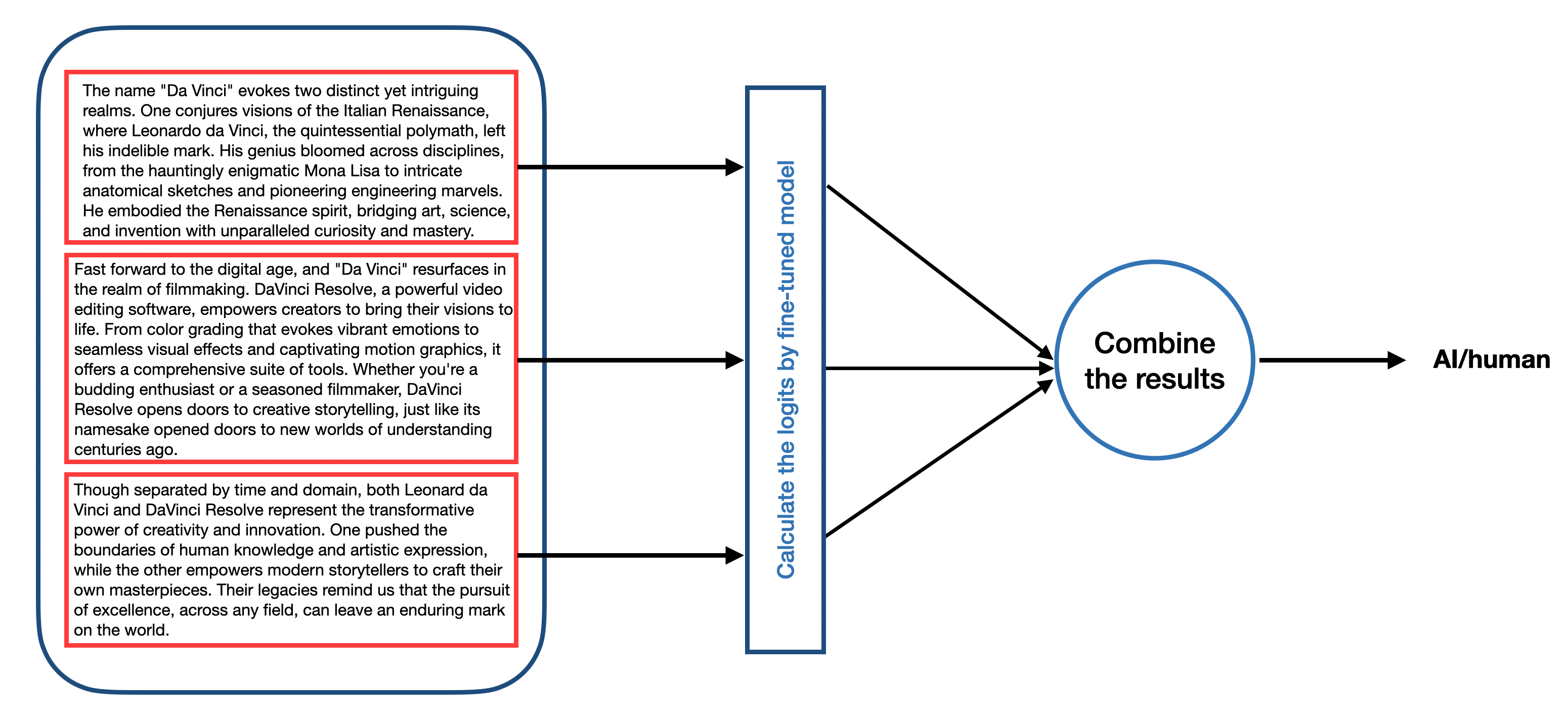}
    \caption{The input text is divided into meaningful units, and the probability of each segment based on their logits is assessed using a fine-tuned XLM-RoBERTa model; the combined evidence leads to a definitive classification.}
    \label{fig1}
\centering
\end{figure*}

\section{Background}

Today, due to the remarkable advancements in Natural Language Processing (NLP), models like ChatGPT, GPT-3, Gemini (formerly known as Bard), and others have reached a point where they can generate texts that closely resemble human writing. Consequently, the task of identifying texts generated by AI has become increasingly important. This task holds significant value across various domains, including content moderation, plagiarism detection, and ensuring transparency in AI-generated content. The approaches for this task can be categorized into three categories: (1) Deep Learning-based Detection, (2) Statistical Discrepancy Detection, and (3) Watermark-based Detection. Deep learning-based models can be formulated as a classification task where the input is a text that can be generated by either a human or an AI. The model is trained with labeled data, where each text is assigned a label indicating whether it was generated by AI or by a human. This allows the model to learn patterns and features that can accurately classify texts based on their origin. These methods are susceptible to adversarial attacks, which can manipulate the input text to deceive the model's classification. However, deep learning-based models generally demonstrate good performance on the training data distribution \cite{guo2023how}. Statistical Discrepancy Detection methods first learn the patterns of AI-generated and human-written texts separately. Then, they identify statistical discrepancies between these patterns to distinguish between the two. By analyzing various linguistic features, such as word frequencies, sentence structures, or syntactic patterns, these methods can detect differences that arise from the distinct nature of AI-generated and human-written texts. Some tools like GPTZero \cite{mitchell2023detectgpt} use perplexity (how well a language model predicts the next word based on the previous ones) and burstiness (variations in sentence length) to assess whether the text is AI-generated or human-written. The idea of watermarking initially emerged from the field of computer vision and has since been applied to NLP \cite{wu2023survey}. This method involves embedding a hidden "watermark" during the text generation process with the objective of identifying text generated by a specific language model. In the context of black-box language models, \cite{yang2023watermarking} utilize this watermarking method to detect and identify text generated by such models.

In this work, we have used the dataset provided by the M4 article, which is a multilingual dataset. The human-written texts in this dataset were collected from diverse sources spanning different domains. These sources include Wikipedia (March 2022 version), WikiHow, Reddit (ELI5), arXiv, and PeerRead for English. For Chinese, the texts were sourced from Baike and Web question answering (QA). Additionally, texts from news sources were included for Urdu and Indonesian, while for Russian, texts were obtained from RuATD. For Arabic, the texts were collected from Arabic Wikipedia. For the monolingual section, we have used the English corpora. In this dataset, AI-generated texts leverage multilingual LLMs such as ChatGPT, textdavinci-003, LLaMa, FlanT5, Cohere, Dolly-v2, and BLOOMz. These models undertake diverse tasks, including creating Wikipedia articles from titles and abstracts (from arXiv), generating peer reviews from titles and abstracts (PeerRead), answering questions from platforms like Reddit and Baike/Web QA, and composing news briefs based on the title. This dataset contains 122k human–machine parallel data in total, with 101k for English, 9k for Chinese, 9k for Russian, 9k for Urdu, 9k for Indonesian, and 9k for Arabic, respectively \cite{wang2023m4}.

For our experiment, we used the English corpora of this dataset in the monolingual track. For the multilingual track, we utilized the whole dataset, which contains human-written and AI-generated texts from six different languages: English, Arabic, Chinese, Indonesian, Russian, and Urdu. As it is evident, our model's input is text documents, and its output is a single label that specifies whether the given text is human-written or AI-generated.


\section{Method}
To classify texts as either AI-generated or human-written, we have examined two crucial aspects: semantics and syntax. Our analysis of these aspects, which is detailed below, aims to identify distinctive features.

    \subsection{Semantic Approach}
    In our exploration of the semantic aspects of texts, we centered our analysis on two key elements: the vocabulary choices employed by the writer and the manner in which words are structured and combined. To achieve this, we leveraged transformers, which utilize word embeddings to capture meaning and positional encoding to account for word order and sentence structure. However, a significant challenge lies in differentiating between AI-generated and human-written texts, especially in longer pieces, as AI models become increasingly adept at mimicking human writing styles. To address this challenge, particularly in longer texts, we adopted a strategy of splitting the text into smaller paragraphs. This allowed for a more focused and detailed analysis of each individual segment, potentially revealing subtle semantic nuances that might be overlooked in a holistic approach.
    
    Our methodology is implemented in three distinct stages: (1) text segmentation, where the input text is divided into meaningful units; (2) probability calculation, where the likelihood of each segment being AI-generated or human-written is assessed; and (3) final prediction, where the combined evidence leads to a definitive classification. A visual representation of our approach is shown in Figure \ref{fig1}.
    
    In the first stage, the input text was segmented into smaller units by splitting it at points where specific markers, such as exclamation marks, question marks, and periods, appeared within paragraphs. Additionally, during this stage, a dataset was generated to fine-tune our model.
    
    For the second stage, we fine-tuned an XLM-RoBERTa model \cite{conneau2019unsupervised} on the aforementioned dataset. Due to limited resources and constraints, the model was trained for only three epochs with a learning rate of \verb|10^-8|.
    After the text was segmented and the model was fine-tuned, we proceeded to analyze the characteristics of each segment and calculate the probability of it being AI-generated or human-written according to their logits. To determine the final results, we employed several methods to combine the results, which are outlined below:
    
    \begin{itemize}
        \item  \textbf{Soft voting prediction:} In this approach, we calculate the average probability of segments; if the calculated average is higher than the threshold (0.95), we conclude that the text is AI-generated. 
        \item \textbf{Hard voting prediction:} In this approach, we calculate the probability of each segment; if it is higher than the threshold, we consider it to be AI-generated, and if more than half of the segments are considered AI-generated, then we conclude that the text is also AI-generated.
        \item \textbf{Weighted soft voting:} This approach is like soft voting, but we give weight to each segment; the weight of each segment is based on the number of words it contains.
        
    \end{itemize}

We use this 0.95 threshold because of the small number of epochs the model has been trained on data.
    
    \subsection{Syntactic Approach}
    Another aspect that we examined was the syntactic properties of the texts. To analyze these properties, we utilized the Part-of-Speech (POS) labels associated with the words in the text. Our goal was to classify AI-generated and human-written texts based on their POS patterns.
    
    To create a dataset for training a model on this aspect, we employed Trankit\cite{van2021trankit}, which provided us with the Universal POS (UPOS) tokens. We integrated these UPOS tokens to form sequences of UPOS strings. In this approach, the focus is on identifying patterns within the sequences rather than the specific meaning of the tokens.
    
    Given the challenge of working with long sequences, we opted to use an LSTM model. To handle the complexity of the task, we employed stacked LSTM layers. Additionally, to enhance the model's performance, we utilized bidirectional LSTMs, which consider the context from both directions of the sequence.
    In order to further improve the model's ability to capture important syntactic patterns and dependencies, we incorporated an attention layer into our LSTM model. The attention mechanism allows the model to focus on specific parts of the input sequence when making predictions, assigning different weights to different elements in the sequence.

    \begin{figure}
    \centering
    \includegraphics[width=8cm]{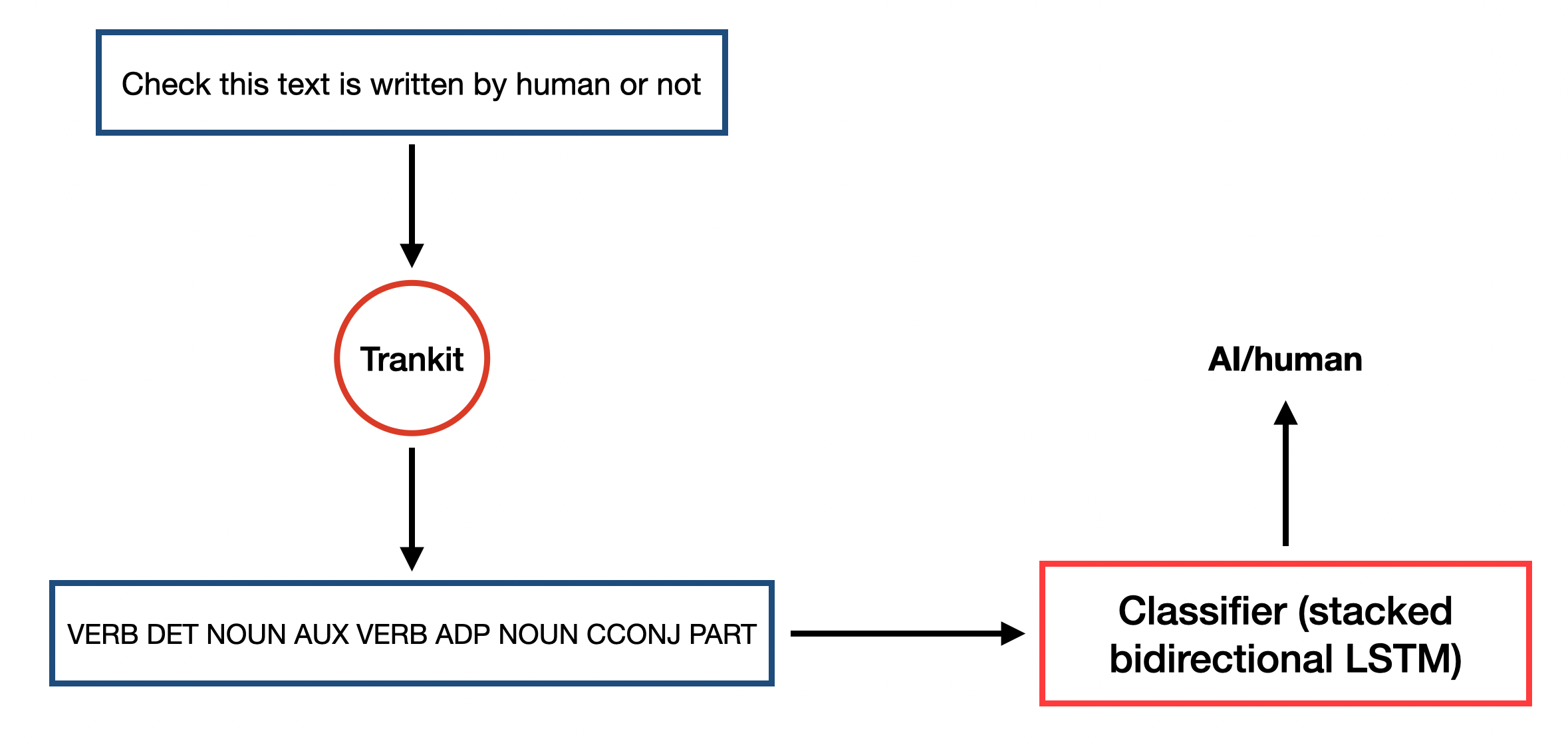}
    \caption{The bidirectional LSTM model predicts using part-of-speech labels associated with the words in the text assessed by Trankit.}
    \label{fig3}
    \end{figure}
    
    By using LSTM models instead of transformer models, we aimed to prevent the potential effects of semantic meaning from overshadowing the syntactic patterns. This choice allowed us to place emphasis on the structural aspects of the texts and better isolate the syntactic features for classification purposes.
    
    By combining the LSTM architecture with an attention layer, we aimed to enhance the model's ability to capture and utilize the important syntactic patterns in the text, ultimately improving the accuracy and effectiveness of the classification process.
    However, the results indicate that there is no specific difference between AI-generated and human-written texts in terms of their UPOS (Universal Part-of-Speech) patterns. We attained an accuracy of 49.75\% and an F1 score of 33\% for both the micro and macro averages. Therefore, we only considered the semantic aspects and overlooked the syntactic aspects.


\section{Result}
    
Our classifier secured 17th rank in multilingual and 27th rank in monolingual competition.

The results for our model are shown in Table~\ref{tab:metrics}. Our model achieved 0.847 and 0.859 accuracy for multilingual and monolingual test datasets, respectively, by training only on the multilingual training dataset.

\begin{figure*}[ht]
    \centering
        \frame{\includegraphics[width=16cm]{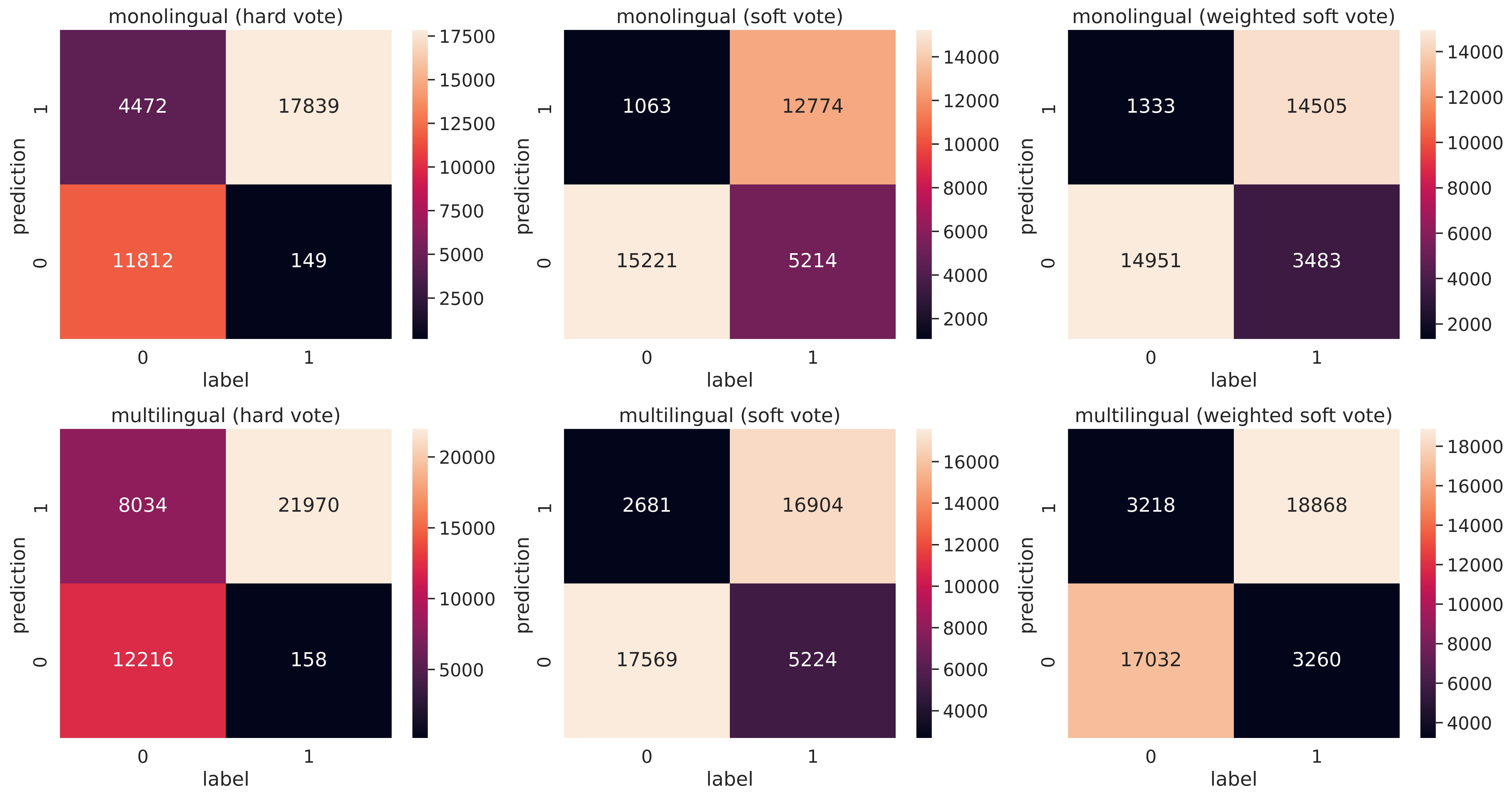}}
        \caption{Confusion matrices of our model for test datasets on monolingual and multilingual tracks}
    {\label{fig:confusion-matrices}}
\end{figure*}

As you can see in the confusion matrix plots in Figure~\ref{fig:confusion-matrices},  weighted soft vote performs better than soft, and soft vote performs better than hard vote. The hard vote approach has higher false positive error than the other two; the soft vote approach has slightly less false positive error but has much more false negative error than the weighted soft vote approach.

By taking a look at mispredicted samples, we realized that the model is weak in predicting formal texts, like texts about history, law, or academic topics.

\begin{table}[ht]
\centering
\begin{tabular}{lll}
\hline
\textbf{metric} & \textbf{multilingual} & \textbf{monolingual}\\
\hline
accuracy & {0.847} & {0.859} \\
precision & {0.854} & {0.916} \\
recall & {0.853} & {0.806} \\
f1 & {0.853} & {0.858} \\
false positive rate & {0.159} & {0.082} \\
false negative rate & {0.147} & {0.194} \\
\hline
\end{tabular}
\caption{
Performance of our classifier according to official metrics
}
{\label{tab:metrics}}
\end{table}

\section{Conclusion}
In this study, we proposed a system to distinguish between human-generated and AI-generated texts. Our approach considered both semantic and syntactic aspects. For the semantic analysis, we focused on smaller text segments instead of the entire document, as we believed that AI models could produce similarly coherent long texts as humans. The results confirmed our assumption.

Our syntactic analysis, which employed a basic model to categorize texts based on their grammatical patterns using UPOS tags, revealed no significant differences in UPOS tag distribution between AI-generated and human-written texts. However, the analysis of word order identified distinct patterns in the semantic approach. This finding suggests that relying solely on UPOS tags for differentiation may be insufficient.

In conclusion, our proposed system demonstrated superior performance compared to the official baseline, achieving a 3.9\% improvement in the multilingual subtask. These results emphasize the significance of considering texts in smaller segments rather than analyzing them as a whole. Moreover, our discoveries suggest that focusing solely on grammar, as indicated by their UPOS tags, may not sufficiently distinguish between AI-generated and human-written texts. Therefore, for future research efforts, it might be beneficial to utilize Graph Neural Networks (GNNs) to examine the grammatical connections among words. This involves representing word embeddings as nodes and their grammatical relationships as edges based on their constituency parsing or dependency parsing trees.

\bibliography{acl_latex}

\end{document}